\documentclass[letterpaper]{article}

\usepackage{aaai}
\usepackage{times} 

\newcommand{\citet}[1]{\citeauthor{#1}~\shortcite{#1}}
\newcommand{\citep}{\cite}

\usepackage{ifxetex}

\ifxetex
 \usepackage{fontspec}
\else
 \usepackage[T1]{fontenc}
 \usepackage[utf8]{inputenc}
\fi

\usepackage[utf8]{inputenc} 
\usepackage[T1]{fontenc}    
\usepackage{url}            
\usepackage{booktabs}       
\usepackage{amsfonts}       
\usepackage{nicefrac}       
\usepackage{microtype}      
\usepackage{makecell}       
\usepackage{graphicx}       

\usepackage{caption}

\usepackage{amsmath,amssymb}
\usepackage{tabularx}
\usepackage{lscape}

\newcommand{\norm}[1]{\left\lVert#1\right\rVert}
\newcommand{\bigO}[1]{\mathcal{O}(#1)}
\newcommand{\emb}[1]{\ensuremath{\mathbf{e}_{#1}}}
\newcommand{\remb}[1]{\ensuremath{\mathbf{r}_{#1}}}
\newcommand{\tdot}[3]{\ensuremath{\langle #1, #2, #3 \rangle}}

\DeclareMathOperator{\vect}{vec}

\newcommand{\ents}{\ensuremath{\mathcal{E}}}
\newcommand{\rels}{\ensuremath{\mathcal{R}}}

\newcommand{\Real}{\ensuremath{\mathbb{R}}}

\newcommand{\Complex}{\ensuremath{\mathbb{C}}}

\newcommand{\fs}{\ensuremath{\psi}}

\usepackage{graphicx}
\usepackage{float}

\usepackage{xargs}
\usepackage[pdftex,dvipsnames]{xcolor}

\usepackage{titlesec}
\usepackage[titletoc,toc,title]{appendix}

\title{Convolutional 2D Knowledge Graph Embeddings}

\author{
    Tim Dettmers\thanks{This work was conducted during a research visit to University College London.} \\
    Università della Svizzera italiana \\
    \texttt{tim.dettmers@gmail.com} \\
    \AND
    Pasquale Minervini \quad Pontus Stenetorp \quad Sebastian Riedel\\
    University College London\\
    \texttt{\{p.minervini,p.stenetorp,s.riedel\}@cs.ucl.ac.uk} \\
}

\begin{document}

\maketitle

\begin{abstract}
Link prediction for knowledge graphs is the task of predicting missing relationships between entities.
Previous work on link prediction has focused on shallow, fast models which can scale to large knowledge graphs.
However, these models learn less expressive features than deep, multi-layer models -- which potentially limits performance.
In this work we introduce ConvE, a multi-layer convolutional network model for link prediction, and report state-of-the-art results for several established datasets.
We also show that the model is highly parameter efficient, yielding the same performance as DistMult and R-GCN with 8x and 17x fewer parameters.
Analysis of our model suggests that it is particularly effective at modelling nodes with high indegree -- which are common in highly-connected, complex knowledge graphs such as Freebase and YAGO3.
In addition, it has been noted that the WN18 and FB15k datasets suffer from test set leakage, due to inverse relations from the training set being present in the test set -- however, the extent of this issue has so far not been quantified.
We find this problem to be severe: a simple rule-based model can achieve state-of-the-art results on both WN18 and FB15k.
To ensure that models are evaluated on datasets where simply exploiting inverse relations cannot yield competitive results, we investigate and validate several commonly used datasets -- deriving robust variants where necessary.
We then perform experiments on these robust datasets for our own and several previously proposed models, and find that ConvE achieves state-of-the-art Mean Reciprocal Rank across most datasets.
\end{abstract}

\section{Introduction}

Knowledge graphs are graph-structured knowledge bases, where facts are represented in the form of relationships (edges) between entities (nodes).
They have important applications in search, analytics, recommendation, and data integration -- however, they tend to suffer from incompleteness, that is, missing links in the graph.
For example, in Freebase and DBpedia more than 66\% of the person entries are missing a birthplace~\citep{DBLP:conf/kdd/0001GHHLMSSZ14,DBLP:conf/semweb/KrompassBT15}.
Identifying such missing links is referred to as \emph{link prediction}.
Knowledge graphs can contain millions of facts; as a consequence, link predictors should scale in a manageable way with respect to both the number of parameters and computational costs to be applicable in real-world scenarios.
For solving such scaling problems, link prediction models are often composed of simple operations, like inner products and matrix multiplications over an embedding space, and use a limited number of parameters~\citep{nickel2016review}.
DistMult~\citep{yang15:embedding} is such a model, characterised by three-way interactions between embedding parameters, which produce one feature per parameter.
Using such simple, fast, shallow models allows one to scale to large knowledge graphs, at the cost of learning less expressive features.
The only way to increase the number of features in shallow models -- and thus their expressiveness -- is to increase the embedding size.
However, doing so does not scale to larger knowledge graphs, since the total number of embedding parameters is proportional to the the number of entities and relations in the graph.
For example, a shallow model like DistMult with an embedding size of 200, applied to Freebase, will need 33~GB of memory for its parameters.
To increase the number of features independently of the embedding size requires the use of multiple layers of features.
However, previous multi-layer knowledge graph embedding architectures, that feature fully connected layers, are prone to overfit~\citep{nickel2016review}.
One way to solve the scaling problem of shallow architectures, and the overfitting problem of fully connected deep architectures, is to use parameter efficient, fast operators which can be composed into deep networks.
The convolution operator, commonly used in computer vision, has exactly these properties: it is parameter efficient and fast to compute, due to highly optimised GPU implementations.
Furthermore, due to its ubiquitous use, robust methodologies have been established to control overfitting when training multi-layer convolutional networks~\citep{szegedy2015going,ioffe2015batch,srivastava2014dropout,szegedy2016rethinking}.
In this paper we introduce ConvE, a model that uses 2D convolutions over embeddings to predict missing links in knowledge graphs.
ConvE is the simplest multi-layer convolutional architecture for link prediction: it is defined by a single convolution layer, a projection layer to the embedding dimension, and an inner product layer.
Specifically, our contributions are as follows:

\begin{itemize}
 \item{
    Introducing a simple, competitive 2D convolutional link prediction model, ConvE.
}
 \item{
    Developing a 1-N scoring procedure that speeds up training three-fold and evaluation by 300x.
}
 \item{
    Establishing that our model is highly parameter efficient, achieving better scores than DistMult and R-GCNs on FB15k-237 with 8x and 17x fewer parameters.
}
\item{
    Showing that for increasingly complex knowledge graphs, as measured by indegree and PageRank, the difference in performance between our model and a shallow model increases proportionally to the complexity of the graph.
}
 \item{
    Systematically investigating reported inverse relations test set leakage across commonly used link prediction datasets, introducing robust versions of datasets where necessary, so that they cannot be solved using simple rule-based models.
}
\item{
    Evaluating ConvE and several previously proposed models on these robust datasets: our model achieves state-of-the-art Mean Reciprocal Rank across most of them.
}
\end{itemize}
\section{Related Work}
Several neural link prediction models have been proposed in the literature, such as the Translating Embeddings model (TransE)~\citep{DBLP:conf/nips/BordesUGWY13}, the Bilinear Diagonal model (DistMult)~\citep{yang15:embedding} and its extension in the complex space (ComplEx)~\citep{DBLP:conf/icml/TrouillonWRGB16}; we refer to \citet{nickel2016review} for a recent survey.
The model that is most closely related to this work is most likely the Holographic Embeddings model (HolE)~\citep{DBLP:conf/aaai/NickelRP16}, which uses cross-correlation -- the inverse of circular convolution -- for matching entity embeddings; it is inspired by holographic models of associative memory.
However, HolE does not learn multiple layers of non-linear features, and it is thus theoretically less expressive than our model.
To the best of our knowledge, our model is the first neural link prediction model to use 2D convolutional layers.
\emph{Graph Convolutional Networks}~(GCNs)~\citep{DBLP:conf/nips/DuvenaudMABHAA15,DBLP:conf/nips/DefferrardBV16,DBLP:journals/corr/KipfW16} are a related line of research, where the convolution operator is generalised to use locality information in graphs.
However, the GCN framework is limited to undirected graphs, while knowledge graphs are naturally directed, and suffers from potentially prohibitive memory requirements~\citep{DBLP:journals/corr/KipfW16}.
Relational GCNs (R-GCNs)~\citep{schlichtkrull2017modeling} are a generalisation of GCNs developed for dealing with highly multi-relational data such as knowledge graphs -- we include them in our experimental evaluations.
Several convolutional models have been proposed in natural language processing (NLP) for solving a variety of tasks, including semantic parsing~\citep{DBLP:conf/conll/YihTPM11}, sentence classification~\citep{DBLP:conf/emnlp/Kim14}, search query retrieval~\citep{DBLP:conf/www/ShenHGDM14}, sentence modelling~\citep{DBLP:conf/acl/KalchbrennerGB14}, as well as other NLP tasks~\citep{DBLP:journals/jmlr/CollobertWBKKK11}.
However, most work in NLP uses 1D-convolutions, that is convolutions which operate over a temporal sequence of embeddings, for example a sequence of words in embedding space.
In this work, we use 2D-convolutions which operate spatially over embeddings.
\subsection{Number of Interactions for 1D vs 2D Convolutions}
Using 2D rather than 1D convolutions increases the expressiveness of our model through additional points of interaction between embeddings.
For example, consider the case where we concatenate two rows of 1D embeddings, $a$ and $b$ with dimension $n=3$: 
\begin{equation*}
\left(
   \begin{bmatrix}
   a & a & a
\end{bmatrix} 
;
    \begin{bmatrix}
b &b & b
 \end{bmatrix} 
 \right)
 = 
    \begin{bmatrix}
 a& a & a&b&b & b
 \end{bmatrix}.
\end{equation*}
A padded 1D convolution with filter size $k = 3$ will be able to model the interactions between these two embeddings around the concatenation point (with a number of interactions proportional to $k$).
If we concatenate (i.e. stack) two rows of 2D embeddings with dimension $m \times n$, where $m=2$ and $n=3$, we obtain the following:
\begin{equation*}
\left(
   \begin{bmatrix}
   a & a  & a \\ a & a & a
\end{bmatrix} 
;
    \begin{bmatrix}
b &b & b\\ b & b & b
 \end{bmatrix} 
 \right)
 = 
    \begin{bmatrix}
 a & a & a\\ a & a & a\\ b & b & b\\ b & b & b
 \end{bmatrix}.
\end{equation*}
A padded 2D convolution with filter size $3 \times 3$ will be able to model the interactions around the entire concatenation line (with a number of interactions proportional to $n$ and $k$).
We can extend this principle to an alternating pattern, such as the following:
\begin{equation*}
    \begin{bmatrix}
 a & a & a\\ b & b & b\\ a & a & a\\ b & b & b
 \end{bmatrix}.
 \end{equation*}
In this case, a 2D convolution operation is able to model even more interactions between $a$ and $b$ (with a number of interactions proportional to $m$, $n$, and $k$).
Thus, 2D convolution is able to extract more feature interactions between two embeddings compared to 1D convolution.
The same principle can be extending to higher dimensional convolutions, but we leave this as future work.
\section{Background}
\begin{center}
\begin{table*}[ht]
\caption{Scoring functions $\fs_{r}(\emb{s}, \emb{o})$ from neural link predictors in the literature, their relation-dependent parameters and space complexity; $n_e$ and $n_r$ respectively denote the number of entities and relation types, i.e. $n_e = |\ents|$ and $n_r = |\rels|$.} \label{tab:overview}
\begin{center}
\begin{tabularx}{\textwidth}{lccc}
\toprule
\multicolumn{1}{c}{\bf Model} & {\bf Scoring Function} $\fs_{r}(\emb{s}, \emb{o})$ & {\bf Relation Parameters} & {\bf Space Complexity} \\
\hline
%
%
SE~\citep{DBLP:journals/ml/BordesGWB14} & $\norm{\mathbf{W}^{L}_{r} \emb{s} - \mathbf{W}^{R}_{r} \emb{o}}_{p}$ & $\mathbf{W}^{L}_{r}, \mathbf{W}^{R}_{r} \in \Real^{k \times k}$ & $\bigO{n_{e} k + n_{r} k^{2}}$\\
TransE~\citep{DBLP:conf/nips/BordesUGWY13} & $\norm{\emb{s} + \remb{r} - \emb{o}}_{p}$ & $\remb{r} \in \Real^{k}$ & $\bigO{n_{e} k + n_{r} k}$ \\
DistMult~\citep{yang15:embedding} & $\tdot{\emb{s}}{\remb{r}}{\emb{o}}$ & $\remb{r} \in \Real^{k}$ & $\bigO{n_{e} k + n_{r} k}$ \\
ComplEx~\citep{DBLP:conf/icml/TrouillonWRGB16} & $\tdot{\emb{s}}{\remb{r}}{\emb{o}}$ & $\remb{r} \in \Complex^{k}$ & $\bigO{n_{e} k + n_{r} k}$ \\
ConvE & $f ( \vect ( f ([ \overline{\emb{s}} ; \overline{\remb{r}} ] \ast \omega )) \mathbf{W} )\emb{o}$ & $\remb{r} \in \Real^{k'}$ & $\bigO{n_{e} k + n_{r} k'}$ \\
\bottomrule
\end{tabularx}
\end{center}
\end{table*}
\end{center}
A \emph{knowledge graph} $\mathcal{G} = \{ (s, r, o) \} \subseteq \ents \times \rels \times \ents$ can be formalised as a set of triples (facts), each consisting of a relationship $r \in \rels$ and two entities $s, o \in \ents$, referred to as the \emph{subject} and \emph{object} of the triple.
Each triple $(s, r, o)$ denotes a relationship of type $r$ between the entities $s$ and $o$.
The \emph{link prediction} problem can be formalised as a pointwise learning to rank problem, where the objective is learning a scoring function $\fs : \ents \times \rels \times \ents \mapsto \Real$.
Given an input triple $x = (s, r, o)$, its score $\fs(x) \in \Real$ is proportional to the likelihood that the fact encoded by $x$ is true.
\subsection{Neural Link Predictors}
Neural link prediction models~\citep{nickel2016review} can be seen as multi-layer neural networks consisting of an \emph{encoding component} and a \emph{scoring component}.
Given an input triple $(s, r, o)$, the encoding component maps entities $s, o \in \ents$ to their distributed embedding representations $\emb{s}, \emb{o} \in \Real^{k}$.
In the scoring component, the two entity embeddings $\emb{s}$ and $\emb{o}$ are scored by a function $\fs_{r}$.
The score of a triple $(s, r, o)$ is defined as $\fs(s, r, o) = \fs_{r}(\emb{s}, \emb{o}) \in \Real$.
In Table~\ref{tab:overview} we summarise the scoring function of several link prediction models from the literature.
The vectors $\emb{s}$ and $\emb{o}$ denote the subject and object embedding, where $\emb{s}, \emb{o} \in \Complex^{k}$ in ComplEx and $\emb{s}, \emb{o} \in \Real^{k}$ in all other models, and $\tdot{x}{y}{z} = \sum_{i} x_{i} y_{i} z_{i}$ denotes the tri-linear dot product; $\ast$ denotes the convolution operator; $f$ denotes a non-linear function.

\section{Convolutional 2D Knowledge Graphs Embeddings}

\begin{figure*}[ht]
	\caption{In the ConvE model, the entity and relation embeddings are first reshaped and concatenated (steps 1, 2); the resulting matrix is then used as input to a convolutional layer (step 3); the resulting feature map tensor is vectorised and projected into a $k$-dimensional space (step 4) and matched with all candidate object embeddings (step 5).}
	\label{fig:model}
	\includegraphics[width=1.0\linewidth]{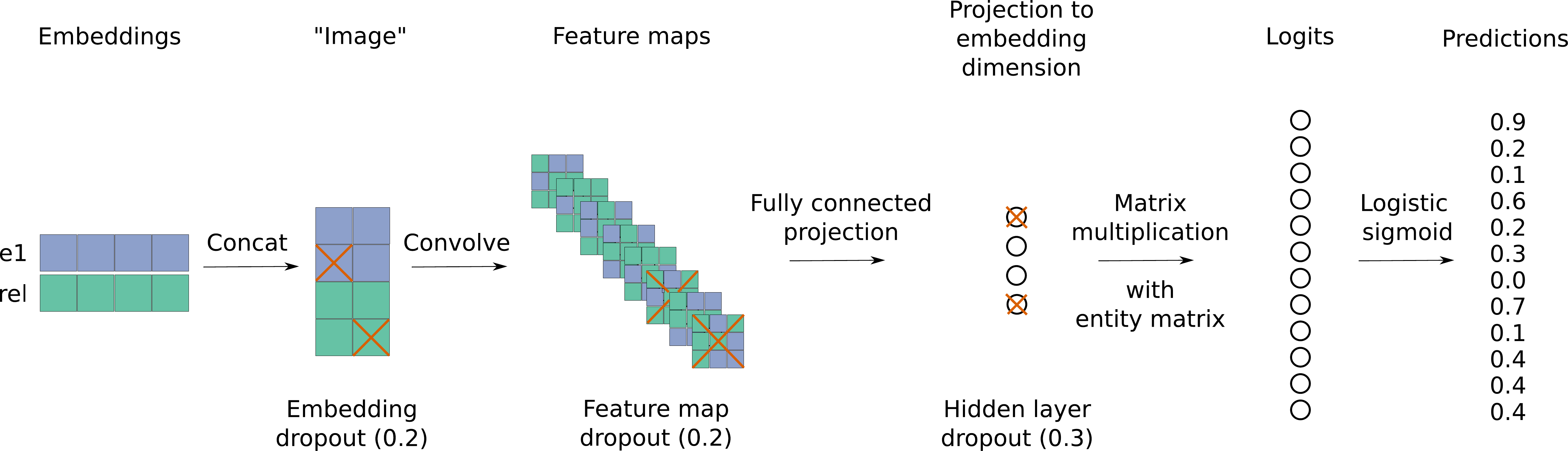}
\end{figure*}

In this work we propose a neural link prediction model where the interactions between input entities and relationships are modelled by convolutional and fully-connected layers.
The main characteristic of our model is that the score is defined by a convolution over 2D shaped embeddings.
The architecture is summarised in Figure~\ref{fig:model}; formally, the scoring function is defined as follows:
\begin{equation}
\begin{aligned}
 \fs_{r}(\emb{s}, \emb{o}) & = & f ( \vect ( f ( [ \overline{\emb{s}} ; \overline{\remb{r}} ] \ast \omega ) ) \mathbf{W} ) \emb{o}, \\
\end{aligned}
\end{equation}
\noindent where $\remb{r} \in \Real^{k}$ is a relation parameter depending on $r$, $\overline{\emb{s}}$ and $\overline{\remb{r}}$ denote a 2D reshaping of $\emb{s}$ and $\remb{r}$, respectively: if $\emb{s}, \remb{r} \in \Real^{k}$, then $\overline{\emb{s}}, \overline{\remb{r}} \in \Real^{k_{w} \times k_{h}}$, where $k = k_{w} k_{h}$.

In the feed-forward pass, the model performs a row-vector look-up operation on two embedding matrices, one for entities, denoted ${\bf E}^{|\ents| \times k}$ and one for relations, denoted ${ \bf R}^{|\rels| \times k'}$, where $k$ and $k'$ are the entity and relation embedding dimensions, and $|\ents|$ and $|\rels|$ denote the number of entities and relations.
The model then concatenates $\overline{\emb{s}}$ and $\overline{\remb{r}}$, and uses it as an input for a 2D convolutional layer with filters $\omega$.
Such a layer returns a feature map tensor $\mathcal{T} \in \Real^{c \times m \times n}$, where $c$ is the number of 2D feature maps with dimensions $m$ and $n$.
The tensor $\mathcal{T}$ is then reshaped into a vector $\vect(\mathcal{T}) \in \Real^{c m n}$, which is then projected into a $k$-dimensional space using a linear transformation parametrised by the matrix $\mathbf{W} \in \Real^{c m n \times k}$ and matched with the object embedding $\emb{o}$ via an inner product.
The parameters of the convolutional filters and the matrix $\mathbf{W}$ are independent of the parameters for the entities $s$ and $o$ and the relationship $r$.

For training the model parameters, we apply the logistic sigmoid function $\sigma(\cdot)$ to the scores, that is $p = \sigma(\fs_{r}(\emb{s}, \emb{o}))$, and minimise the following binary cross-entropy loss:
\begin{equation}
  \mathcal{L}(p, t) = - \frac{1}{N} \sum_i (t_i \cdot \log(p_i) + (1 - t_i) \cdot \log(1 - p_i)),
\end{equation}
where $t$ is the label vector with dimension $\Real^{1x1}$ for 1-1 scoring or $\Real^{1xN}$ for 1-N scoring (see the next section for 1-N scoring); the elements of vector $t$ are ones for relationships that exists and zero otherwise.

We use rectified linear units as the non-linearity $f$ for faster training~\citep{krizhevsky2012imagenet}, and batch normalisation after each layer to stabilise, regularise and increase rate of convergence~\citep{ioffe2015batch}.
We regularise our model by using dropout~\citep{srivastava2014dropout} in several stages.
In particular, we use dropout on the embeddings, on the feature maps after the convolution operation, and on the hidden units after the fully connected layer.
We use Adam as optimiser~\citep{kingma2014adam}, and label smoothing to lessen overfitting due to saturation of output non-linearities at the labels~\citep{szegedy2016rethinking}.

\subsection{Fast Evaluation for Link Prediction Tasks}
\label{onetoN}

In our architecture convolution consumes about 75-90\% of the total computation time, thus it is important to minimise the number of convolution operations to speed up computation as much as possible.
For link prediction models, the batch size is usually increased to speed up evaluation~\citep{bordes2013translating}.
However, this is not feasible for convolutional models since the memory requirements quickly outgrow the GPU memory capacity when increasing the batch size.

Unlike other link prediction models which take an entity pair and a relation as a triple $(s, r, o)$, and score it (1-1 scoring), we take one $(s, r)$ pair and score it against all entities $o \in \ents$ simultaneously (1-N scoring).
If we benchmark 1-1 scoring on a high-end GPU with batch size and embedding size 128, then a training pass and an evaluation with a convolution model on FB15k -- one of the dataset used in the experiments -- takes 2.4 minutes and 3.34 hours.
Using 1-N scoring, the respective numbers are 45 and 35 seconds -- a considerable improvement of over 300x in terms of evaluation time.
Additionally, this approach is scalable to large knowledge graphs and increases convergence speed.
For a single forward-backward pass with batch size of 128, going from $N=100,000$ to $N=1,000,000$ entities only increases the computational time from 64ms to 80ms -- in other words, a ten-fold increase in the number of entities only increases the computation time by 25\% -- which attests the scalability of the approach. 

If instead of 1-N scoring, we use 1-(0.1N) scoring -- that is, scoring against 10\% of the entities -- we can compute a forward-backward pass 25\% faster.
However, we converge roughly 230\% slower on the training set.
Thus 1-N scoring has an additional effect which is akin to batch normalisation~\citep{ioffe2015batch} -- we trade some computational performance for greatly increased convergence speed and also achieve better performance as shown in Section~\ref{ablation_study}.
Do note that the technique in general could by applied to any 1-1 scoring model.
This practical trick in speeding up training and evaluation can be applied to any 1-1 scoring model, such as the great majority of link prediction models.

\section{Experiments}
\label{Experiments}
\subsection{Knowledge Graph Datasets}

For evaluating our proposed model, we use a selection of link prediction datasets from the literature.
WN18~\citep{DBLP:conf/nips/BordesUGWY13} is a subset of WordNet which consists of 18 relations and 40,943 entities.
Most of the 151,442 triples consist of hyponym and hypernym relations and, for such a reason, WN18 tends to follow a strictly hierarchical structure. 
FB15k~\citep{DBLP:conf/nips/BordesUGWY13} is a subset of Freebase which contains about 14,951 entities with 1,345 different relations.
A large fraction of content in this knowledge graph describes facts about movies, actors, awards, sports, and sport teams. 
YAGO3-10~\citep{DBLP:conf/cidr/MahdisoltaniBS15} is a subset of YAGO3 which consists of entities which have a minimum of 10 relations each.
It has 123,182 entities and 37 relations.
Most of the triples deal with descriptive attributes of people, such as citizenship, gender, and profession.
Countries~\citep{bouchard2015approximate} is a benchmark dataset that is useful to evaluate a model's ability to learn long-range dependencies between entities and relations.
It consists of three sub-tasks which increase in difficulty in a step-wise fashion, where the minimum path-length to find a solution increases from 2 to 4.
It was first noted by~\citet{toutanova2015observed} that WN18 and FB15k suffer from test leakage through inverse relations: a large number of test triples can be obtained simply by inverting triples in the training set.
For example, the test set frequently contains triples such as $(s, \text{hyponym}, o)$ while the training set contains its inverse $(o, \text{hypernym}, s)$.
To create a dataset without this property, \citet{toutanova2015observed} introduced FB15k-237 -- a subset of FB15k where inverse relations are removed.
However, they did not explicitly investigate the severity of this problem, which might explain why research continues using these datasets for evaluation without addressing this issue (e.g. \citet{DBLP:conf/icml/TrouillonWRGB16}, \citet{DBLP:conf/aaai/NickelRP16}, \citet{nguyen2016stranse}, \citet{liu2016hierarchical}).

In the following section, we introduce a simple rule-based model which demonstrates the severity of this bias by achieving state-of-the-art results on both WN18 and FB15k.
In order to ensure that we evaluate on datasets that do not have inverse relation test leakage, we apply our simple rule-based model to each dataset.
Apart from FB15k, which was corrected by FB15k-237, we also find flaws with WN18.
We thus create WN18RR to reclaim WN18 as a dataset, which cannot easily be completed using a single rule -- but requires modelling of the complete knowledge graph.
WN18RR\footnote{\url{https://github.com/TimDettmers/ConvE}} contains 93,003 triples with 40,943 entities and 11 relations.
For future research, we recommend against using FB15k and WN18 and instead recommend FB15k-237, WN18RR, and YAGO3-10.
\subsection{Experimental Setup}
We selected the hyperparameters of our ConvE model via grid search according to the mean reciprocal rank (MRR) on the validation set.
Hyperparameter ranges for the grid search were as follows -- embedding dropout $\{ 0.0, 0.1, 0.2 \}$, feature map dropout $\{ 0.0, 0.1, 0.2, 0.3 \}$, projection layer dropout $\{ 0.0, 0.1, 0.3, 0.5 \}$, embedding size $\{ 100, 200 \}$, batch size $\{ 64, 128, 256 \}$, learning rate $\{ 0.001, 0.003 \}$, and label smoothing $\{ 0.0, 0.1, 0.2, 0.3 \}$.
Besides the grid search, we investigated modifications of the 2D convolution layer for our models.
In particular, we experimented with replacing it with fully connected layers and 1D convolution; however, these modifications consistently reduced the predictive accuracy of the model.
We also experimented with different filter sizes, and found that we only receive good results if the first convolutional layer uses small (i.e. 3x3) filters.
\begin{table}[t]
	\caption{Parameter scaling of DistMult vs ConvE.}
	\label{parameters}
	\begin{center}
	\begin{tabular}{lcccccc}
	    \toprule
		 & Param. & Emb. & & \multicolumn{3}{c}{{ Hits}}   \\
		Model & count & size  &  MRR & @10 & @3 & @1 \\
		\midrule
		DistMult & 1.89M & 128 &  .23 & .41 & .25 & .15  \\
		DistMult & 0.95M & 64 &  .22 & .39 & .25 & .14  \\
		DistMult & 0.23M & 16 &  .16 & .31 & .17 & .09  \\
		\midrule
		ConvE & 5.05M & 200 &  .32 & .49 & .35 & .23  \\
		ConvE & 1.89M & 96 &  .32 & .49 & .35 & .23  \\
		ConvE & 0.95M & 54 &  .30 & .46 & .33 & .22  \\
		ConvE & 0.46M & 28 &  .28 & .43 & .30 & .20  \\
		ConvE & 0.23M & 14 &  .26 & .40 & .28 & .19  \\
	\bottomrule
\end{tabular}
\end{center}
\end{table}
We found that the following combination of parameters works well on WN18, YAGO3-10 and FB15k: embedding dropout 0.2, feature map dropout 0.2, projection layer dropout 0.3, embedding size 200, batch size 128, learning rate 0.001, and label smoothing 0.1.
For the Countries dataset, we increase embedding dropout to 0.3, hidden dropout to 0.5, and set label smoothing to 0.
We use early stopping according to the mean reciprocal rank (WN18, FB15k, YAGO3-10) and AUC-PR (Countries) statistics on the validation set, which we evaluate every three epochs.
Unlike the other datasets, for Countries the results have a high variance, as such we average 10 runs and produce 95\% confidence intervals. 
For our DistMult and ComplEx results with 1-1 training, we use an embedding size of 100, AdaGrad~\citep{DBLP:journals/jmlr/DuchiHS11} for optimisation, and we regularise our model by forcing the entity embeddings to have a L2 norm of 1 after each parameter update.
As in \citet{DBLP:conf/nips/BordesUGWY13}, we use a pairwise margin-based ranking loss.
The code for our model and experiments is made publicly available,\footnote{\url{https://github.com/TimDettmers/ConvE}} as well as the code for replicating the DistMult results.\footnote{\url{https://github.com/uclmr/inferbeddings}}
\subsection{Inverse Model}
It has been noted by \citet{toutanova2015observed}, that the training datasets of WN18 and FB15k have 94\% and 81\% test leakage as inverse relations, that is, 94\% and 81\% of the triples in these datasets have inverse relations which are linked to the test set.
For instance, a test triple \textit{(feline, hyponym, cat)} can easily be mapped to a training triple \textit{(cat, hypernym, feline)} if it is known that hyponym is the inverse of hypernym.
This is highly problematic, because link predictors that do well on these datasets may simply learn which relations that are the inverse of others, rather than to model the actual knowledge graph.
To gauge the severity of this problem, we construct a simple, rule-based model that solely models inverse relations. We call this model the \emph{inverse model}.
The model extracts inverse relationships automatically from the training set: given two relation pairs $r_1, r_2 \in \rels$, we check whether $(s, r_1, o)$ implies $(o, r_2, s)$, or vice-versa. 
We assume that inverse relations are randomly distributed among the training, validation and test sets and, as such, we expect the number of inverse relations to be proportional to the size of the training set compared to the total dataset size.
Thus, we detect inverse relations if the presence of $(s, r_1, o)$ co-occurs with the presence of $(o, r_2, s)$ with a frequency of at least $0.99 - (f_v + f_t)$, where $f_v$ and $f_t$ is the fraction of the validation and test set compared to the total size of the dataset.
Relations matching this criterion are assumed to be the inverse of each other.
At test time, we check if the test triple has inverse matches outside the test set: if $k$ matches are found, we sample a permutation of the top $k$ ranks for these matches; if no match is found, we select a random rank for the test triple.
\section{Results}

\begin{table*}[t]
	\caption{Link prediction results for WN18 and FB15k}
	\label{results_dirty}
	\centering
	\begin{tabularx}{\textwidth}{lccccccccccc}
	    \toprule
		 & \multicolumn{5}{c}{{ \bf WN18}} & & \multicolumn{5}{c}{{\bf FB15k}} \\
		 	\cmidrule{2-6}
    	\cmidrule{8-12}
    	 & & & \multicolumn{3}{c}{Hits} & & & & \multicolumn{3}{c}{Hits} \\
		& MR & MRR & @10 & @3 & @1 & & MR & MRR & @10 & @3 & @1  \\
		\midrule
		DistMult~\citep{yang15:embedding} & 902 & .822 & .936 & .914 & .728 & & 97 & .654 & .824 & .733 & .546 \\
		ComplEx~\citep{DBLP:conf/icml/TrouillonWRGB16} & -- & .941 & .947  & .936 &  .936 & & -- & .692 & .840 & .759 & .599 \\
		Gaifman~\citep{DBLP:conf/nips/Niepert16} & \textbf{352} & -- & .939 & -- & .761 & & 75 & -- & .842 & -- & \textbf{.692} \\
		ANALOGY~\citep{2017arXiv170502426L} & -- &  .942  & .947 & .944 & .939 & & -- & \textbf{.725} & \textbf{.854} & \textbf{.785} & .646 \\
		R-GCN~\citep{schlichtkrull2017modeling} & -- & .814 & .964 & .929 & .697 & & -- & .696 & .842 & .760 & .601 \\
		\midrule
		ConvE & 374 & .943 & .956 & .946 & .935 & & {\bf 51} & .657 & .831 & .723 & .558 \\		
		Inverse Model & 740 & \textbf{.963} & \textbf{.964} & \textbf{.964} & \textbf{.953} & & 2501 & .660 & .660 & .659 & .658 \\
		\bottomrule
	\end{tabularx}
\end{table*}

\begin{table*}[t]
	\caption{Link prediction results for WN18RR and FB15k-237}
	\label{results_clean}
	\centering	
	\begin{tabularx}{\textwidth}{lccccccccccc}
	    \toprule
		 & \multicolumn{5}{c}{{ \bf WN18RR}} & &  \multicolumn{5}{c}{{ \bf FB15k-237}}                  \\
		\cmidrule{2-6}
    	\cmidrule{8-12}    
    	 & & & \multicolumn{3}{c}{Hits} & & & & \multicolumn{3}{c}{Hits}  \\
    	\cmidrule{4-6}     	\cmidrule{10-12}         
		& MR & MRR & @10 & @3 & @1 & & MR & MRR & @10 & @3 & @1  \\
		\midrule
		DistMult~\citep{yang15:embedding} & {5110} & .43 & .49 & .44 & .39 & & 254 & .241 & .419 & .263 & .155 \\		
	    ComplEx~\citep{DBLP:conf/icml/TrouillonWRGB16} & 5261 & .44 & {.51} & \textbf{.46} & \textbf{.41} & & 339 & .247 & .428 & .275 & .158 \\
	    
	    R-GCN~\citep{schlichtkrull2017modeling}  & -- & -- &  --  & -- & -- & & -- & .248 & .417 & .258  & .153 \\
		\midrule
		ConvE & \textbf{4187} &  {.43} & \textbf{.52} & .44 &  .40 & & \textbf{244} & {\bf.325} & {\bf.501} & {\bf.356} & \textbf{.237} \\
		Inverse Model & 13526 &  .35 & {.35} & {.35} &  .35 & & 7030 & .010 & .014 & .011 & {.007} \\
		\bottomrule
	\end{tabularx}
\end{table*}

\begin{table*}[t]
	\caption{Link prediction results for YAGO3-10 and Countries}
	\label{results_yago}
	\centering
	\begin{tabularx}{\textwidth}{lccccccccc}
	    \toprule
		 & \multicolumn{5}{c}{{ \bf YAGO3-10}}    &  \multicolumn{4}{c}{{ \bf Countries}}                \\
		\cmidrule{2-6} 	\cmidrule{8-10}
		& & & \multicolumn{3}{c}{Hits}    &  \multicolumn{4}{c}{AUC-PR}              \\
		\cmidrule{4-6}  \cmidrule{8-10}
		& MR & MRR & @10 & @3 & @1 & & S1 & S2 & S3  \\
		\midrule
		DistMult~\citep{yang15:embedding} & {5926} & .34 & .54 & .38 & .24 &  & \textbf{1.00\text{$\pm$}0.00}  & 0.72\text{$\pm$}0.12  & 0.52\text{$\pm$}0.07 \\
		 ComplEx~\citep{DBLP:conf/icml/TrouillonWRGB16} & 6351 & {.36} &  {.55}  & .40 & .26  & & 0.97\text{$\pm$}0.02  & 0.57\text{$\pm$}0.10  & 0.43\text{$\pm$}0.07 \\
		\midrule
		ConvE & \textbf{1676} &  \textbf{.44} & \textbf{.62} & \textbf{.49} &  \textbf{.35}  & & \textbf{1.00\text{$\pm$}0.00}  & \textbf{0.99\text{$\pm$}0.01}  & \textbf{0.86 \text{$\pm$}0.05}  \\
		Inverse Model & 59448 &  .01 & {.02} & {.02} &  .01 & & -- & -- & -- \\
	  	\bottomrule
	\end{tabularx}
\end{table*}

Similarly to previous work~\citep{yang15:embedding,DBLP:conf/icml/TrouillonWRGB16,DBLP:conf/nips/Niepert16}, we report results using a filtered setting, i.e. we rank test triples against all other candidate triples not appearing in the training, validation, or test set~\citep{DBLP:conf/nips/BordesUGWY13}.
Candidates are obtained by permuting either the subject or the object of a test triple with all entities in the knowledge graph.
Our results on the standard benchmarks FB15k and WN18 are shown in Table~\ref{results_dirty}; results on the datasets with inverse relations removed are shown in Table~\ref{results_clean}; results on YAGO3-10 and Countries are shown in Table~\ref{results_yago}.
Strikingly, the inverse model achieves state-of-the-art on many different metrics for both FB15k and WN18.
However, it fails to pick up on inverse relations for YAGO3-10 and FB15k-237.
The procedure used by \citet{toutanova2015observed} to derive FB15k-237 does not remove certain symmetric relationships, for example ``similar to''.
The presence of these relationships explains the good score of our inverse model on WN18RR, which was derived using the same procedure.

Our proposed model, ConvE, achieves state-of-the-art performance for all metrics on YAGO3-10, for some metrics on FB15k, and it does well on WN18.
On Countries, it solves the S1 and S2 tasks, and does well on S3, scoring better than other models like DistMult and ComplEx.

For FB15k-237, we could not replicate the basic model results from \citet{toutanova2015representing}, where the models in general have better performance than what we can achieve.
Compared to \citet{schlichtkrull2017modeling}, our results for standard models are a slightly better then theirs, and on-a-par with their R-GCN model.
\subsection{Parameter efficiency of ConvE}
From Table~\ref{parameters} we can see that ConvE for FB15k-237 with 0.23M parameters performs better than DistMult with 1.89M parameters for 3 metrics out of 5.
ConvE with 0.46M parameters still achieves state-of-the-art results on FB15k-237 with 0.425 Hits@10.
Comparing to the previous best model, R-GCN~\citep{schlichtkrull2017modeling}, which achieves 0.417 Hits@10 with more than 8M parameters.
Overall, ConvE is more than 17x parameter efficient than R-GCNs, and 8x more parameter efficient than DistMult.
For the entirety of Freebase, the size of these models would be more than 82GB for R-GCNs, 21GB for DistMult, compared to 5.2GB for ConvE.

\section{Analysis}

\subsection{Ablation Study}

Table~\ref{ablation_study} shows the results from our ablation study where we evaluate different parameter initialisation ($n=2$) to calculate confidence intervals.
We see that hidden dropout is by far the most important component, which is unsurprising since it is our main regularisation technique.
1-N scoring improves performance, as does input dropout, feature map dropout has a minor effect, while label smoothing seems to be unimportant -- as good results can be achieved without it.

\begin{center}
\begin{table}[ht]
	\caption{Mean PageRank $\times$10$^{-3}$ of nodes in the test set vs reduction in error in terms of AUC-PR or Hits@10 of ConvE wrt. DistMult.}
	\label{testset_pagerank}
	\begin{center}
    	\begin{tabular}{lcc}
    	    \toprule       
    		Dataset & PageRank & Error Reduction \\
    		\midrule	    
            WN18RR & 0.104 & 0.06    \\
            WN18 & 0.125 & 0.45    \\
            FB15k & 0.599 & 0.04    \\
            FB15-237 & 0.733 & 0.16    \\
            YAGO3-10 & 0.988 & 0.21    \\
            Countries S3 & 1.415 & 2.36    \\
            Countries S1 & 1.711 & 0.00    \\
            Countries S2 & 1.796 & 17.6    \\
    		\bottomrule
    	\end{tabular}
	\end{center}
\end{table}
\end{center}

\begin{center}
\begin{table}[t]
	\caption{Ablation study for FB15k-237.}
\label{ablation_study}
\begin{center}
\begin{tabular}{lc}
    \toprule       
	Ablation & Hits@10 \\
	\midrule	    
    Full ConvE & 0.491	\\
	\midrule	    
    Hidden dropout & -0.044 $\pm$ 0.003\\
    Input dropout & -0.022 $\pm$ 0.000	\\
    1-N scoring & -0.019	\\
    Feature map dropout & -0.013 $\pm$ 0.001	\\
    Label smoothing & -0.008 $\pm$ 0.000	\\
	\bottomrule
\end{tabular}
\end{center}
\end{table}
\end{center}

\subsection{Analysis of Indegree and PageRank}

Our main hypothesis for the good performance of our model on datasets like YAGO3-10 and FB15k-237 compared to WN18RR, is that these datasets contain nodes with very high relation-specific indegree.
For example the node ``United States'' with edges ``was born in'' has an indegree of over 10,000.
Many of these 10,000 nodes will be very different from each other (actors, writers, academics, politicians, business people) and successful modelling of such a high indegree nodes requires capturing all these differences.
Our hypothesis is that deeper models, that is, models that learn multiple layers of features, like ConvE, have an advantage over shallow models, like DistMult, to capture all these constraints.
However, deeper models are more difficult to optimise, so we hypothesise that for datasets with low average relation-specific indegree~(like WN18RR and WN18), a shallow model like DistMult might suffice for accurately representing the structure of the network.

To test our two hypotheses, we take two datasets with low~(low-WN18) and high~(high-FB15k) relation-specific indegree and reverse them into high~(high-WN18) and low~(low-FB15k) relation-specific indegree datasets by deleting low and high indegree nodes. 
We hypothesise that, compared to DistMult, ConvE will always do better on the dataset with high relation-specific indegree, and vice-versa.

Indeed, we find that both hypotheses hold: for low-FB15k we have ConvE 0.586 Hits@10 vs DistMult 0.728 Hits@10; for high-WN18 we have ConvE 0.952 Hits@10 vs DistMult 0.938 Hits@10. This supports our hypothesis that deeper models such as ConvE have an advantage to model more complex graphs (e.g. FB15k and FB15k-237), but that shallow models such as DistMult have an advantage to model less complex graphs (e.g. WN18 WN18RR).
To investigate this further, we look at PageRank, a measure of centrality of a node. PageRank can also be seen as a measure of the recursive indegree of a node: the PageRank value of a node is proportional to the indegree of this node, its neighbours indegrees, its neighbours-neighbours indegrees and so forth scaled relative to all other nodes in the network. 
By this line of reasoning, we also expect ConvE to be better than DistMult on datasets with high average PageRank~(high connectivity graphs), and vice-versa.

To test this hypothesis, we calculate the PageRank for each dataset as a measure of centrality. We find that the most central nodes in WN18 have a PageRank value more than one order of magnitude smaller than the most central nodes in YAGO3-10 and Countries, and about 4 times smaller than the most central nodes in FB15k. When we look at the mean PageRank of nodes contained in the test sets, we find that the difference of performance in terms of Hits@10 between DistMult and ConvE is roughly proportional to the mean test set PageRank, that is, the higher the mean PageRank of the test set nodes the better ConvE does compared to DistMult, and vice-versa. See Table~\ref{testset_pagerank} for these statistics. The correlation between mean test set PageRank and relative error reduction of ConvE compared to DistMult is strong with $r=0.56$. This gives additional evidence that models that are deeper have an advantage when modelling nodes with high (recursive) indegree.

From this evidence we conclude, that the increased performance of our model compared to a standard link predictor, DistMult, can be partially explained due to our it's ability to model nodes with high indegree with greater precision -- which is possibly related to its depth. 

\section{Conclusion and Future Work}

We introduced ConvE, a link prediction model that uses 2D convolution over embeddings and multiple layers of non-linear features to model knowledge graphs.
ConvE uses fewer parameters; it is fast through 1-N scoring; it is expressive through multiple layers of non-linear features; it is robust to overfitting due to batch normalisation and dropout; and achieves state-of-the-art results on several datasets, while still scaling to large knowledge graphs. 
In our analysis, we show that the performance of ConvE compared to a common link predictor, DistMult, can partially be explained by its ability to model nodes with high (recursive) indegree.
Test leakage through inverse relations of WN18 and FB15k was first reported by~\citet{toutanova2015observed}: we investigate the severity of this problem for commonly used datasets by introducing a simple rule-based model, and find that it can achieve state-of-the-art results on WN18 and FB15k.
To ensure robust versions of all investigated datasets exists, we derive WN18RR.
Our model is still shallow compared to convolutional architecture found in computer vision, and future work might deal with convolutional models of increasing depth.
Further work might also look at the interpretation of 2D convolution, or how to enforce large-scale structure in embedding space so to increase the number of interactions between embeddings.

\subsection{Acknowledgements}

We would like to thank Johannes Welbl, Peter Hayes, and Takuma Ebisu for their feedback and helpful discussions related to this work. We thank Takuma Ebisu for pointing out an error in our Inverse Model script -- the corrected results are slightly better for WN18 and slightly worse for FB15k. We thank Victoria Lin for helping us to unroot and fix a bug where the exclusion of triples during inference worked incorrectly -- the changes did not affect the main results in this work, though some results in the appendix changed (UMLS, Nations).
This work was supported by a Marie Curie Career Integration Award, an Allen Distinguished Investigator Award, a Google Europe Scholarship for Students with Disabilities, and the H2020 project SUMMA. 

\bibliographystyle{aaai}
\bibliography{references}

\appendix

\section{\LARGE {\sc Supplemental Material}}

\newcommand{\Test}{\ensuremath{\mathcal{T}}}

\newcommand{\corr}[1]{\ensuremath{\mathcal{C}(#1)}}
\newcommand{\scorr}[1]{\ensuremath{\mathcal{C}^{\text{s}}(#1)}}
\newcommand{\ocorr}[1]{\ensuremath{\mathcal{C}^{\text{o}}(#1)}}

\newcommand{\KG}{\ensuremath{\mathcal{G}}}
\newcommand{\ff}{\ensuremath{f}}
\newcommand{\card}[1]{\ensuremath{\left\vert{#1}\right\vert}}
\newcommand{\Rank}{\text{rank}}
\newcommand{\cN}{\ensuremath{\mathcal{N}}}
\newcommand{\scdot}{\ensuremath{{}\cdot{}}}

\section{Versions}

\begin{itemize}
 \item 2018-07-04: 
     \begin{itemize}
        \item Added new YAGO3-10 results. The new results are worse on most metrics, but state-of-the-art results are retained. 
        \item I was unable to replicate FB15k scores that I initially reported.\footnote{See \url{https://github.com/TimDettmers/ConvE/issues/26} for details.} 
        \item I update the PageRank table and the reported PageRank-error-reduction correlation to reflect the new scores.
        \item I removed the Nations scores in the appendix. The Nations dataset has a high proportion of inverse relationships and is thus not suitable for the use in research. I do not want to encourage its use.
  \end{itemize}
 \item 2018-04-06: Victoria Lin helped us to find and fix issues\footnote{See \url{https://github.com/TimDettmers/ConvE/issues/18} for more information.} with triple masks during evaluation. We report new numbers for UMLS and Nations in the appendix. The results were unchanged on other datasets that we tested thus far (Kinship, WN18, WN18RR, FB15k-237).
    \item 2018-03-28: 
    \begin{itemize}
        \item New numbers for Inverse Model after bugfix by Takuma Ebisu.
        \item New numbers for UMLS/Nations/Kinship datasets in appendix using the most commonly reported test data splits.
    \end{itemize}
    \item 2018-01-07: Extended AAAI camera ready (6/7/8).
    \item 2017-07-08: Missing grant in acknowledgements.
    \item 2017-07-05: Original NIPS submission (6/7/6).
\end{itemize}

\section{Further ConvE results}

\begin{center}
\begin{table}[h]
	\caption{ConvE link prediction results for UMLS, Nations, and Kinship.}
	\label{UMLS}
\begin{center}
	\begin{tabularx}{\columnwidth}{lcccccc}
	    \toprule
    	& & & & \multicolumn{3}{c}{Hits} \\
    	\cmidrule{5-7}
		Dataset & Model & MR & MRR & @10 & @3 & @1  \\
		\midrule
		UMLS & ConvE & 1 & .94 & .99 & .96 & .92 \\		
		Kinship & ConvE & 2 & .83 & .98 & .92 & .74 \\		
		\bottomrule
	\end{tabularx}
\end{center}
\end{table}
\end{center}

\section{Evaluation Metrics}
We now describe the evaluation metrics used for assessing the quality of the models.
Let $\Test = \{ x_{1}, x_{2}, \ldots, x_{\card{\Test}} \}$ denote the test set.
Following \citet{DBLP:conf/nips/BordesUGWY13}, for the $i$-th test triple $x_{i}$ in $\Test$, we generate all its possible corruptions $\scorr{x_{i}}$ (resp. $\ocorr{x_{i}}$) -- obtained by replacing its subject (resp. object) with any other entity in the Knowledge Graph -- to check whether the model assigns an higher score to $x_{i}$ and a lower score to its corruptions.
Note that the set of corruptions can also contain several true triples, and it is not a mistake to rank them with an higher score than $x_{i}$.
For such a reason, we remove all triples in the graph from the set of corruptions: this is referred to as the \emph{filtered} setting in \citet{DBLP:conf/nips/BordesUGWY13}.
The \emph{left and right rank} of the $i$-th test triple -- each associated to corrupting either the subject or the object -- according to a model with scoring function $\fs(\scdot)$, are defined as follows:
\begin{equation*}
 \begin{aligned}
  \Rank^{\text{s}}_{i} = & 1 + \sum_{\tilde{x}_{i} \in \scorr{x_{i}} \setminus \KG} I\left[\fs(x_{i}) < \fs(\tilde{x}_{i}) \right], \\
  \Rank^{\text{o}}_{i} = & 1 + \sum_{\tilde{x}_{i} \in \ocorr{x_{i}} \setminus \KG} I\left[\fs(x_{i}) < \fs(\tilde{x}_{i}) \right],
 \end{aligned}
\end{equation*}
\noindent where $I\left[P\right]$ is $1$ iff the condition $P$ is true, and $0$ otherwise.
For measuring the quality of the ranking, we use the Mean Reciprocal Rank (MRR) and the Hits@$k$ metrics, which are defined as follows:
\begin{equation*}
 \begin{array}{rl}
  \text{MRR:} & \displaystyle{\frac{1}{2 |\Test|} \sum_{x_{i} \in \Test} \frac{1}{\Rank^{\text{s}}_{i}} + \frac{1}{\Rank^{\text{o}}_{i}} }, \\
  \text{Hits@$k$ (\%):} & \displaystyle{\frac{100}{2 |\Test|} \sum_{x_{i} \in \Test} I\left[ \Rank^{\text{s}}_{i} \leq k \right] + I\left[ \Rank^{\text{o}}_{i} \leq k \right] }.
 \end{array}
\end{equation*}
MRR is the average inverse rank for all test triples: the higher, the better.
Hits@$k$ is the percentage of ranks lower than or equal to $k$: the higher, the better.

\end{document}